\documentclass{llncs}

\usepackage{graphicx}
\usepackage{amsmath}
\usepackage{enumitem}
\usepackage{amsfonts}
\usepackage[linesnumbered]{algorithm2e}
\usepackage{subcaption}
\usepackage{tabularx}
\usepackage{xcolor}
\usepackage{capt-of}
\usepackage{multirow}

\title{Detecting and Explaining Drifts in Yearly Grant Applications} 
\author{Stephen Pauwels \and Toon Calders}
\institute{University of Antwerp\\Antwerp, Belgium\\ ~\\ Universit\'e Libre de Bruxelles\\Brussels, Belgium}

\begin{document}

\maketitle

\begin{abstract}
During the lifetime of a Business Process changes can be made to the workflow, the required resources, required documents, \ldots. Different traces from the same Business Process within a single log file can thus differ substantially due to these  changes. We propose a method that is able to detect concept drift in multivariate log files with a dozen attributes. We test our approach on the BPI Challenge 2018 data consisting of applications for EU direct payment from farmers in Germany where we use it to detect Concept Drift. In contrast to other methods our algorithm does not require the manual selection of the features used to detect drift. Our method first creates a model that captures the relations between attributes and between events of different time steps. This model is then used to score every event and trace. These scores can be used to detect outlying cases and concept drift. Thanks to the decomposability of the score we are able to perform detailed root-cause analysis.  
\end{abstract}

\section{Introduction}
Log files contain valuable information with regard to the execution of different Business Processes in an organization \cite{vanderAalst2011}. By applying different mining techniques on these logs we can extract the workflow, have a better understanding about the resource management, show the differences between departments, \ldots. Most Process Mining techniques require a stable state in order to perform well. Business Processes, however, are prone to different kind of changes, both expected, documented changes as well as unexpected changes may occur. These changes can be induced by new company policies, new regulations, new activities, \ldots. Concept Drift Detection tries to identify the possible change points (I.e. the points in time where changes to the process happen) and tries to explain what caused the drift. After detecting these points it is possible to split the log in smaller logs that do represent the same execution of a Business Process without changes and use standard process mining techniques to further examine the different processes \cite{song2008trace}.

In \cite{pauwels2018extending} we introduced a model that is able to score traces in order to test for anomalies within the log file. In this report we show that the same model can be used to detect drifts within a Process log. Rather than providing a fully-automated solution we offer visual aids that people can use to analyze data. Data that is properly visualized can easily be investigated by people with often better results than both fully-automated systems or fully-manual systems \cite{keim2006challenges}.

We use the BPI Challenge 2018 dataset \cite{dataset2018} to show the capabilities of our drift detection algorithm. The dataset consists of applications for EU direct payments from farmers in Germany. Due to changes in regulations the process changes every year making it a suitable dataset for testing our method.

The goal of this report is to provide visuals that allow us to easily detect concept drifts and helps in the investigation of the main causes of these drift. These visual aids are designed to be easily understood by both business experts and technical people. 

Our report is structured as follows: first we describe existing work in the field of concept drift in Section \ref{sec:related_work}. The dataset itself is described in Section \ref{sec:dataset} together with the few preprocessing steps we have made. In Section \ref{sec:concept_drift} we explain the model used for scoring the events and we introduce a general workflow that uses these scores to determine concept drift. Section \ref{sec:results} shows the results and analysis of the BPIC 2018 data, showing that our method does provide useful tools for detecting concept drift.

\section{Related Work}
\label{sec:related_work}
In Bose et al. \cite{bose2011handling} a concept drift detection method is proposed that uses statistical tests over feature vectors. These feature vectors are based on the order in which activities occur in the workflow. They record for example how many activities always, sometimes, or never occur after a given activity. Other measures can be used to add extra features that may incorporate the other attributes in the log. The method does require manual selection of the exact features to be used in the statistical tests.

Maaradji et al. \cite{maaradji2015fast} propose an automated method for detecting drifts. They focus only on the activities within the process, and transform the traces in the log into partial order runs. These runs summarize the different workflows possible for the process. In order to detect drifts they search for statistically significant changes in the distribution of the most recent runs.

Accorsi et al. \cite{accorsi2011discovering} propose to use clustering of traces to detect drifts in the process. The clustering is done by using the number of intermediate activities between activities within different windows.

Carmona et al. \cite{carmona2012online} create an abstract representation of the process in the form of a polyhedron, which is afterwards used to test how well the representation includes the traces in from log. When the traces differ from the representation a drift is detected. To find other drifts the entire method needs to be repeated. This method is able to work in an online environment where we can check for drifts during the execution of the process, whereas other methods require a log file containing completed traces.

\section{Dataset}
\label{sec:dataset}
The dataset \cite{dataset2018} consists of applications for EU agricultural grants from the last three years. It contains both attributes that were known when the application was submitted and attributes that contain the outcome of the application after the trace finished. We call the former type of attributes the \emph{basic attributes} and the latter type the \emph{derived attributes}.

The basic attributes contain information about the activity, the applicant, his parcels, the document used, etc, whereas the derived attributes contain whether or not the grant was assigned, possible penalties on the requested amount and the actual amount granted.

Unless otherwise stated we only use the basic attributes for building and testing the model, making it possible to create a model that can be used to test traces while they are still going on and detect drift as soon as possible.

The data consists of 2,514,266 events grouped in 43,809 traces each representing a single application. The shortest trace contains only 24 events, the longest 2,973 and on average there are 57 events per trace.

\begin{example}
\emph{A typical snippet of an example trace can be found in Table \ref{tab:example}, which is taken from the BPIC dataset. Some attributes remain the same within a case (I.e. \emph{case}, \emph{year} and \emph{department}) while the others may have different values.}
\begin{table}[t]
	\centering
	\scriptsize
	\begin{tabular}{| c | c | c | c | c | c | c |}
	\hline
	case & activity & doctype & subprocess & success & year & department\\
	\hline
	8b99873a6136cfa6 & mail income & Payment application & Application & true & 2015 & e7 \\
	8b99873a6136cfa6 & mail valid & Payment application & Application & true & 2015 & e7 \\
	8b99873a6136cfa6 & mail valid & Entitlement application & Main & true & 2015 & e7 \\
	8b99873a6136cfa6 & initialize & Parcel document & Main & true & 2015 & e7 \\
	8b99873a6136cfa6 & initialize & Control summary & Main & true & 2015 & e7 \\
	8b99873a6136cfa6 & begin editing & Control summary & Main & true & 2015 & e7 \\
	8b99873a6136cfa6 & finish editing & Control summary & Main & true & 2015 & e7 \\
	$\vdots$ & $\vdots$ & $\vdots$ & $\vdots$ & $\vdots$ & $\vdots$ & $\vdots$\\
	8b99873a6136cfa6 & begin payment & Payment application & Application & true & 2015 & e7 \\
	8b99873a6136cfa6 & finish payment & Payment application & Application & true & 2015 & e7 \\
	\hline
	\end{tabular}
	\caption{Example trace from the dataset}
	\label{tab:example}
\end{table}

\end{example}

\section{Concept Drift Detection}
\label{sec:concept_drift}
In this section we first explain the model we use for scoring the events and introduce a workflow for detecting drifts and finding differences in the data.

\subsection{The Model}
Our model (extended Dynamic Bayesian Network or eDBN) is based on Dynamic Bayesian Networks (DBNs) \cite{russell2009artificial}, an extension of the well-known Bayesian Networks. These networks are capable of capturing Conditional Dependencies between various variables both within one single time step as over different time steps. As we have indicated in \cite{pauwels2018extending}, the DBNs contain useful elements but lack some expressiveness in order to be useful in the context of Business Process logs. We introduce a second type of dependency, the Functional Dependency (FD), that describes mappings of values from one attribute to another.

\begin{example}
\emph{
An example structure based on the dataset for an eDBN is given in Figure \ref{fig:edbn_example}. In the dataset every applicant has the same value for \emph{area} and \emph{young\_farmer} within a single trace. Therefor there exist a Functional Dependency from \emph{applicant} to both \emph{area} and \emph{young\_farmer} in the current time step. And since these values do not change between events in a trace we also link each of these attributes from the previous time step to the current time step.}

\emph{The attribute \emph{activity} in the previous time step has a Conditional Dependency to \emph{activity} in the current time step because the current activity depends on the previous. Branches in the workflow may occur so multiple activities can follow a certain activity, giving the explanation of why this is a CD and not a FD.}

\begin{figure}[t]
\centering
\includegraphics[scale=0.40]{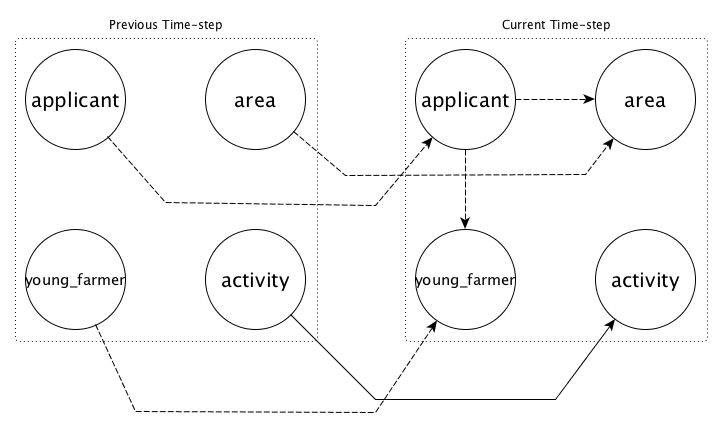}
\caption{eDBN example with conditional (full) and functional dependencies (dashed) based on the dataset.}
\label{fig:edbn_example}
\end{figure}
\end{example}

After finding the different dependencies within the model we have to learn the parameters in order to be able to score events. These parameters consist of all the Conditional Probability Tables, the Functional Dependency Mappings and the probability of encountering new values or new relations as explained in \cite{pauwels2018extending}. We add these \emph{new values} and \emph{new relations} measures to the model to be able to handle new (unseen) values and relations. DBNs return a probability of 0 when encountering a value that was not present in the training dataset, but our log files do allow the occurrence of new values (a new employee entering the company, \ldots) in the logs. Using these measures we can take this into account without severely penalizing the score for an event in case of a justified new value.

\begin{example}
\emph{In order to learn all the parameters for the model in Figure \ref{fig:edbn_example} we start by learning the CPT for the relation \emph{activity\_previous} $\rightarrow$ \emph{activity}. An example CPT is shown in Table \ref{tab:cpt_example}.}
\begin{table}[t]
	\centering
	\begin{tabular}{| l | l | c |}
		\hline
		Parent & Value & Probability\\
		\hline
		\hline
		mail income & mail valid & 1.0 \\
		\hline
		initialize & initialize & 0.6 \\
		& begin editing & 0.3 \\
		& performed & 0.1 \\
		\hline
		begin editing & finish editing & 0.7\\
		& calculate & 0.3\\ 
		\hline
	\end{tabular}
	\caption{CPT for relation between activity\_previous and activity}
	\label{tab:cpt_example}
\end{table}
\emph{For the different FDs we store the mappings that are present between values of the attributes in the training data. The FDT of the relation \emph{applicant} $\rightarrow$ \emph{young\_farmer} will link all applicants to the fact whether they are young farmers or not. In order to determine the \emph{new values} and \emph{new relations} rate, we calculate for every attribute in the current time step the probability of encountering a new value or relation. }
\end{example}

The score for an event $e$ given the model can be calculated by:
\begin{align*}
	Score(e) = \prod_{A \in \{\text{attributes of } e\}}{P(e.A | Parents(A))}
\end{align*}
with
\begin{align*}
P(e.A | Parents(A)) =~&value_A(e.A) \\
				&\times CPT(e.A | Parents(A)) \\
				&\times FDT(e.A)
\end{align*}

Where $value_A$ indicates the score for the value itself (does it occur in the training set or is it a new value), $CPT$ returns the probability for a particular value given the values of the attributes it depends on, and $FDT$ returns a score based on whether or not the Functional Dependencies were broken or not.

The score for an event is built from the contributions of every attribute equally. This helps us to split the score into partial scores were we get the contribution of every attribute by itself. Furthermore the partial score itself can also be decomposed into its components giving an even better and more detailed way of determining the root cause of changes and differences. 

Thanks to the decomposability of the score assigned to an event we created an expressive model that is able to easily give an explanation about how the score was computed. Furthermore thanks to how we composed the score, equal events and traces get approximately the same score making it easy to group traces together that show the same behavior and making it possible to detect drifts and differences.

\subsection{Detecting Drifts}
\begin{figure}[t]
\centering
\includegraphics[scale=0.3]{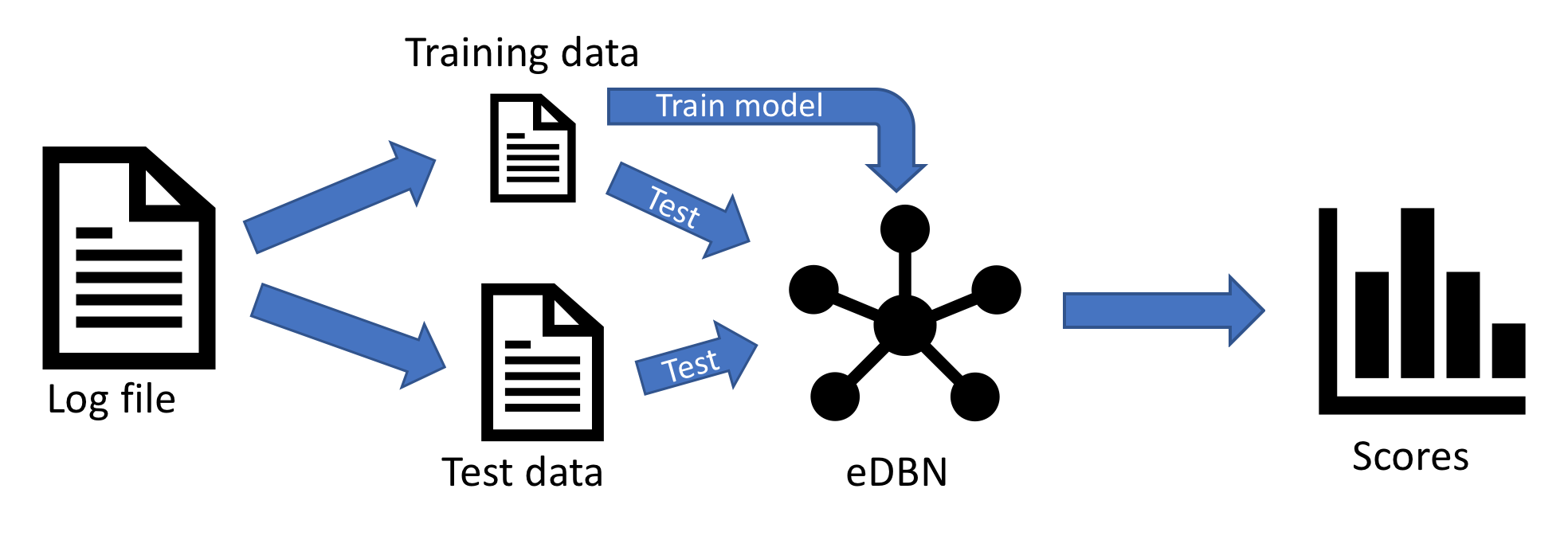}
\caption{Workflow used to score the traces in the log.}
\label{fig:basic-workflow}
\end{figure}

First we select the traces in the log we want to use as the reference set, we call these traces the \emph{training dataset}. This training dataset forms the baseline where we will compare all other traces too, we refer to these traces as the \emph{test dataset}. When detecting concept drift we always select the first $n$ traces as our training dataset and compare them with the other traces. Traces that occur after a drift are supposed to receive a different score than the ones before the drift. The basic workflow of building the model and calculating the scores is given in Figure \ref{fig:basic-workflow}. We can also use this workflow with different kind of training data (E.g. only selecting events that satisfy a certain condition) to find different types of drifts or differences.

In order to be able to compare traces we assign a score to every trace based on the model we created using the training data. This score is calculated as the mean score for every event in the trace. We use this score in contrast to the $n$-th root of the product of all event scores (where $n$ is the number of events in the trace) used in \cite{pauwels2018extending} as we are only looking to characterize a trace using this score. When we would use the original formula to compute the score for a trace, a single violation of a FD can result in the total score being 0. When looking for anomalies this is an important property as these FDs are supposed not to be broken. Also it would be impossible to find out if only one attribute had a partial score of 0 or multiple. By using the mean score we ensure that we can see this difference.

\begin{definition}
The score for a trace is computed as: $Score_{trace}(t) = \frac{\sum_{e \in t}{Score(e)}}{| t |}$. With $| t |$ equal to the number of events in a trace. When there is no confusion possible between the score of an event and a trace we omit the subscript.
\end{definition}

After we have calculated the scores for every trace we can sort them according to timestamp. Traces that represent the same Business Process (without changes in the workflow or attributes) are supposed to have scores closely related to each other. After a drift occurred the scores for the traces change due to the changes that were made to the process itself. The distribution of scores after such a change point is different than before the drift. 
\newpage
\subsubsection{Trace-score plot}
\begin{figure}[t]
\centering
\includegraphics[scale=0.5]{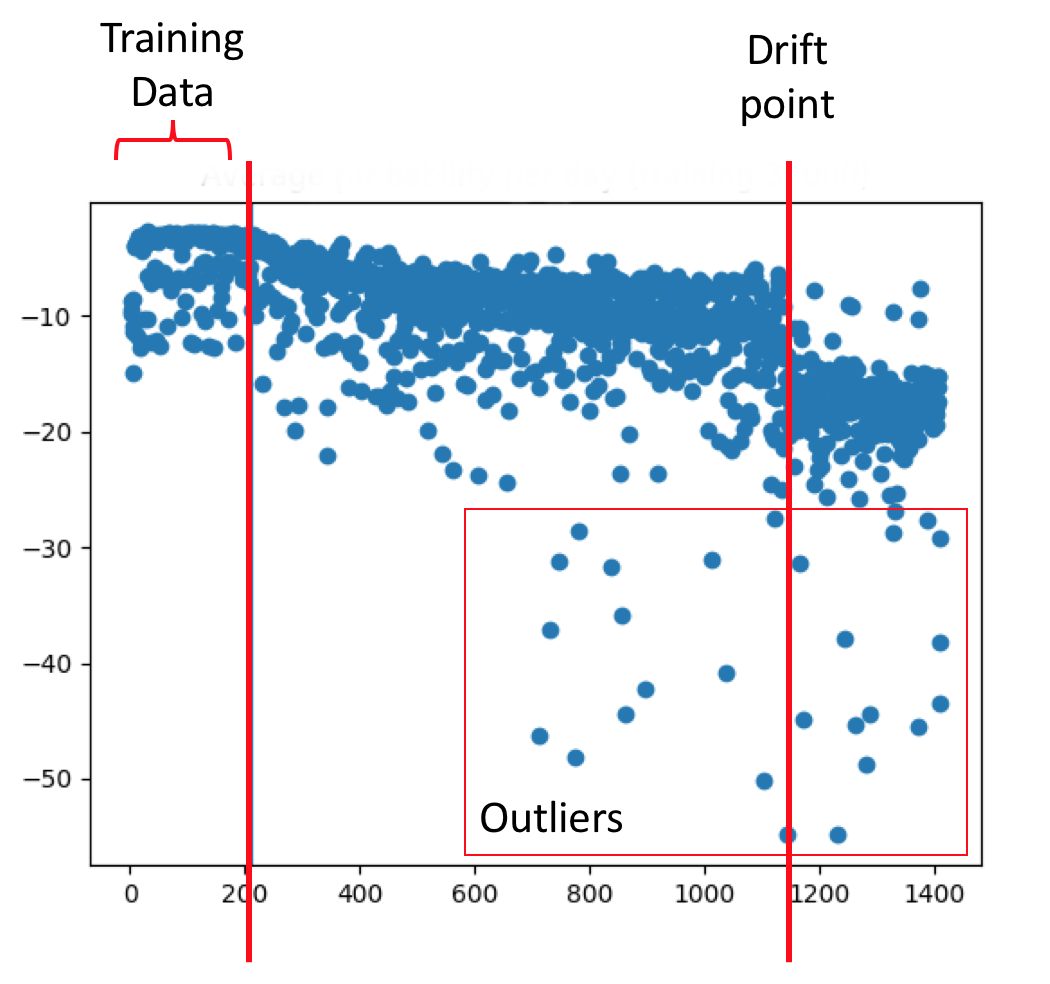}
\caption{Example of scores for different traces picked from a random dataset. The first vertical red line indicates the difference between the traces used for training and testing, the second red line indicates the drift. }
\label{fig:score_example}
\end{figure}

The first visual aid we introduce is the \emph{trace-score plot} which can be seen in Figure \ref{fig:score_example}. The score of every trace is plotted on this graph, giving us a first idea of the different variations present in the log. We use a logarithmic scale for plotting the different scores because the scores of the traces are very small as they are based on probabilities. The logarithmic scale helps us better visualize these differences. The x axis in this plot is time, mostly indicated by the incremental number of the trace in the log. This plot helps us to look at how the scores are distributed over time. A first phenomenon that we can observe is that the overall distribution of scores changes after a certain point in time, these changes represent drifts in the business process. A second interesting observation that can be made is that of outlying traces that deviate from the main distribution of scores. We can then decide to investigate these outlying traces in more detail, as checking all traces in detail is often not feasible.

\newpage
\subsubsection{Drift plot}
\begin{figure}[t]
\centering
\includegraphics[scale=0.4, trim= 0 0 150 0]{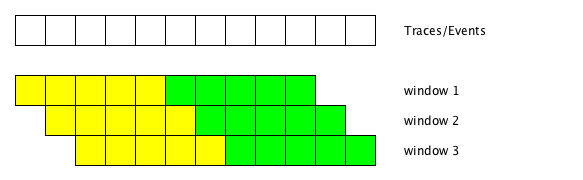}
\caption{Sliding window of size 10 with the \emph{reference window} (yellow) and \emph{test window} (green).}
\label{fig:sliding_window}
\end{figure}

\begin{figure}[h!]
\centering
\includegraphics[scale=0.27]{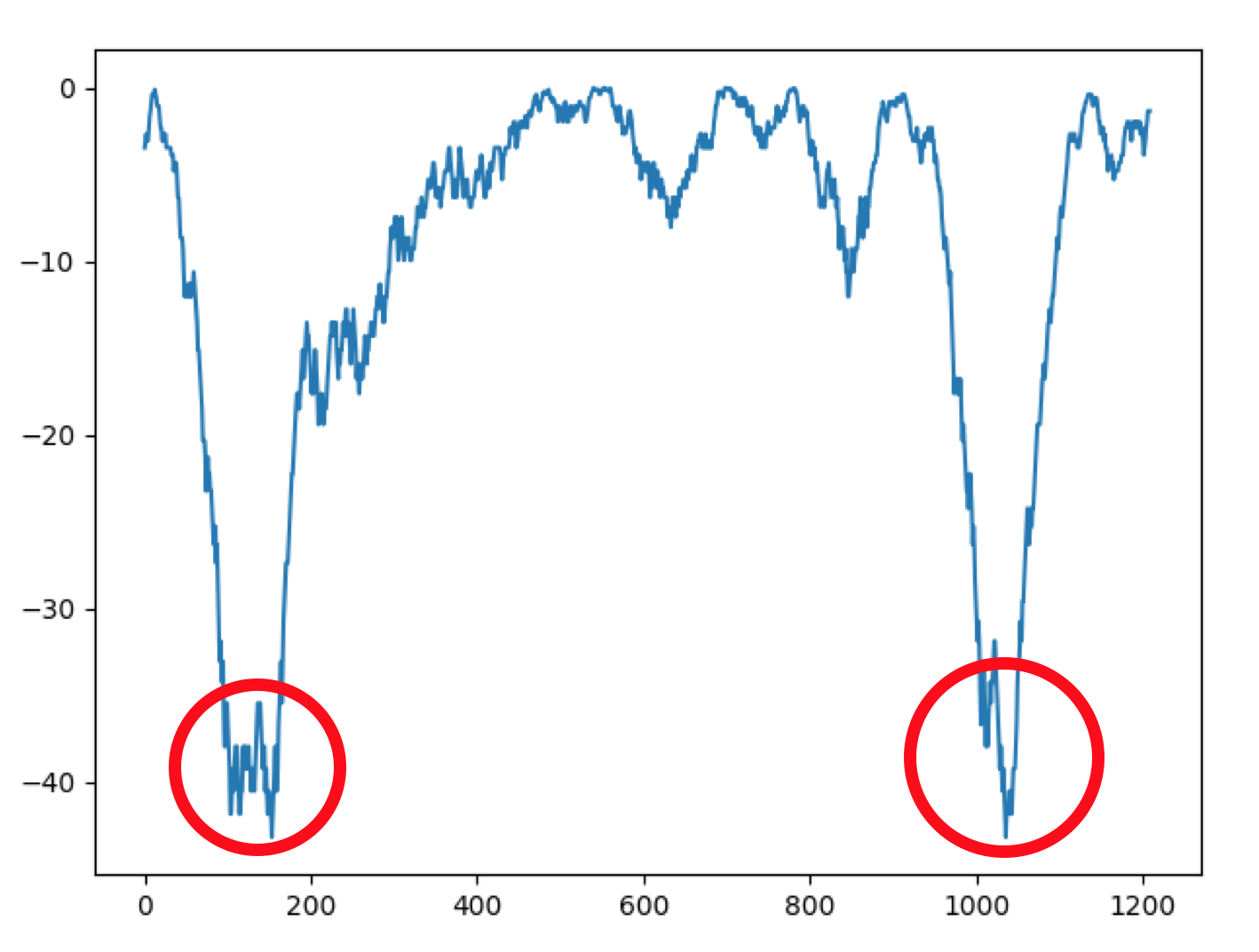}
\caption{Plotting all the p-values obtained by the Kolmogorov-Smirnov test. We can see two dips in this plot. The first dip indicates the change point between the training data and test data. The second dip indicates the real drift point.}
\label{fig:drift_example}
\end{figure}

To better visualize the changes in the scores we use the two sample Kolmogorov-Smirnov statistical test for checking the hypotheses that the distributions in both samples are similar. The test returns a p-value between 0 and 1, where 1 stands for accepting the hypothesis that both distributions are equal and 0 rejecting it. To find the drift points in the entire log we use a sliding window technique. The first half of the window is used as the \emph{reference window} and the second half as the \emph{testing window}, as can be seen in Figure \ref{fig:sliding_window}. We apply the two sample Kolmogorov-Smirnov test on both halves of the window to find possible drift points. The size of the window has to be manually set. A smaller window is more sensitive for finding the smaller drifts but is also more likely to return false positives and a larger window is less sensitive but might return more false negatives. Since we have all the scores calculated and they do not depend on the window size, it is doable to check some different window sizes and compare the plots they return. We can plot all the resulting p-values from the statistical test in order to visually see drift points as shown in Figure \ref{fig:drift_example}. Again we make use of a logarithmic scale since the differences are otherwise not visible. We will refer to this type of plots as the \emph{drift plot}.

\subsubsection{Attribute-density plot}
\begin{figure}[t]
\centering
\includegraphics[scale=0.4]{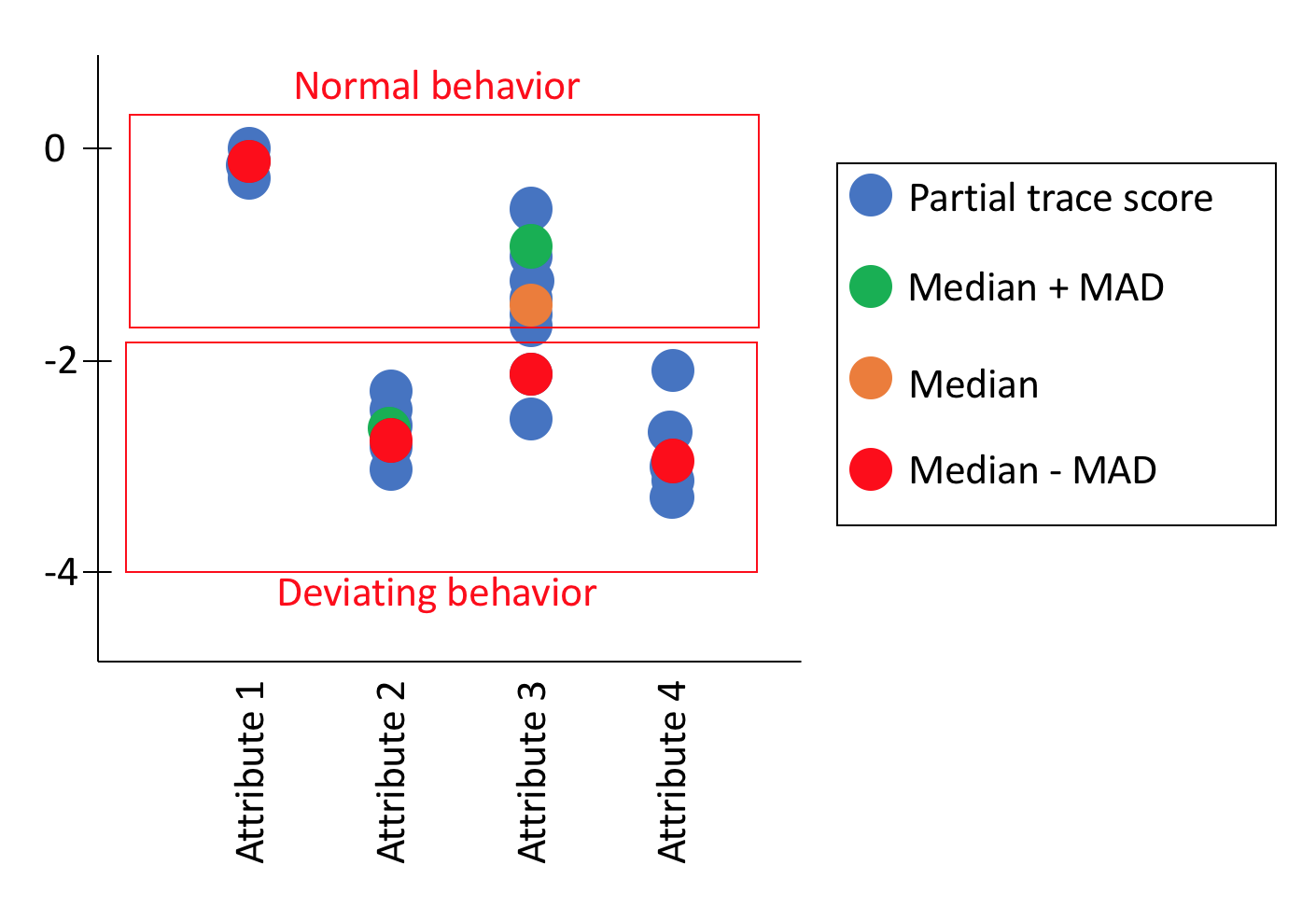}
\caption{Example attribute-density plot for 4 attributes.}
\label{fig:attribute_example}
\end{figure}

To find the root cause for the drifts we found, we take a look at the partial scores of the events, instead of the total event scores themselves. We accumulate the scores according to the attributes. We can plot all the partial trace scores in an \emph{attribute-density plot} as can been seen in Figure \ref{fig:attribute_example}, as before we use a logarithmic scale. We add the median and median absolute deviation \cite{leys2013detecting} in order to get a better insight in the density and distribution of scores for a single attribute. In contrast to the other types of plots these plots characterizes the data in terms of the different attributes. To be able to use these plots for comparing different segments of a log file we first select a reference segment which is used to learn the model used for calculating the partial event scores of all segments. Every segment can be visualized in a different attribute-density plot for comparison. Any technique can be used to help users find differences between these plots. In the remainder of this paper we simply plotted all the median values on the same plot in order to determine differences.

\subsubsection{Towards a general workflow}
\begin{figure}
\centering
\includegraphics[scale=0.21]{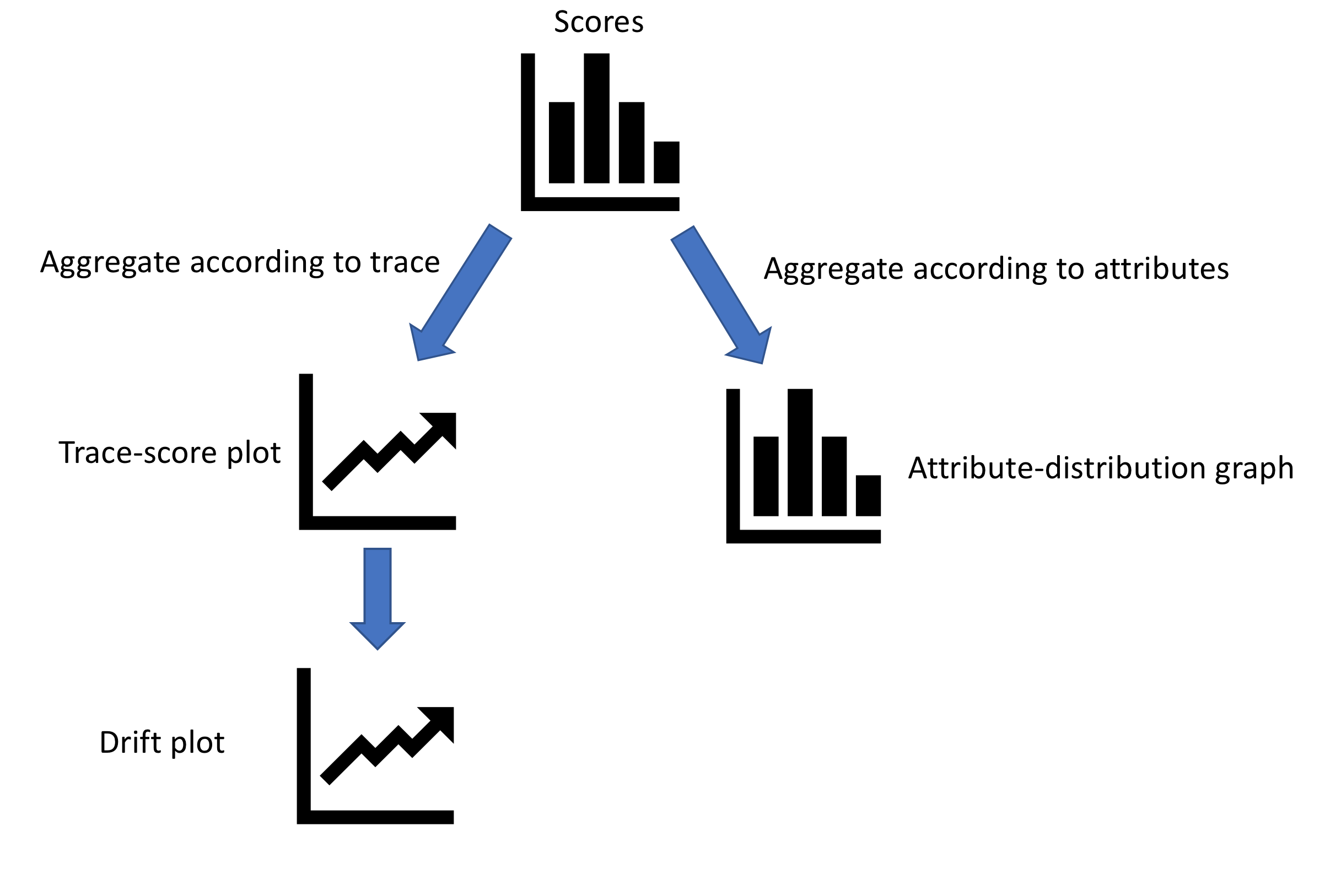}
\caption{Workflow of building the different types of plots and showing how they relate to each other.}
\label{fig:workflow_graphs}
\end{figure}

\begin{figure}
\centering
\includegraphics[scale=0.31, trim = 125 0 0 0]{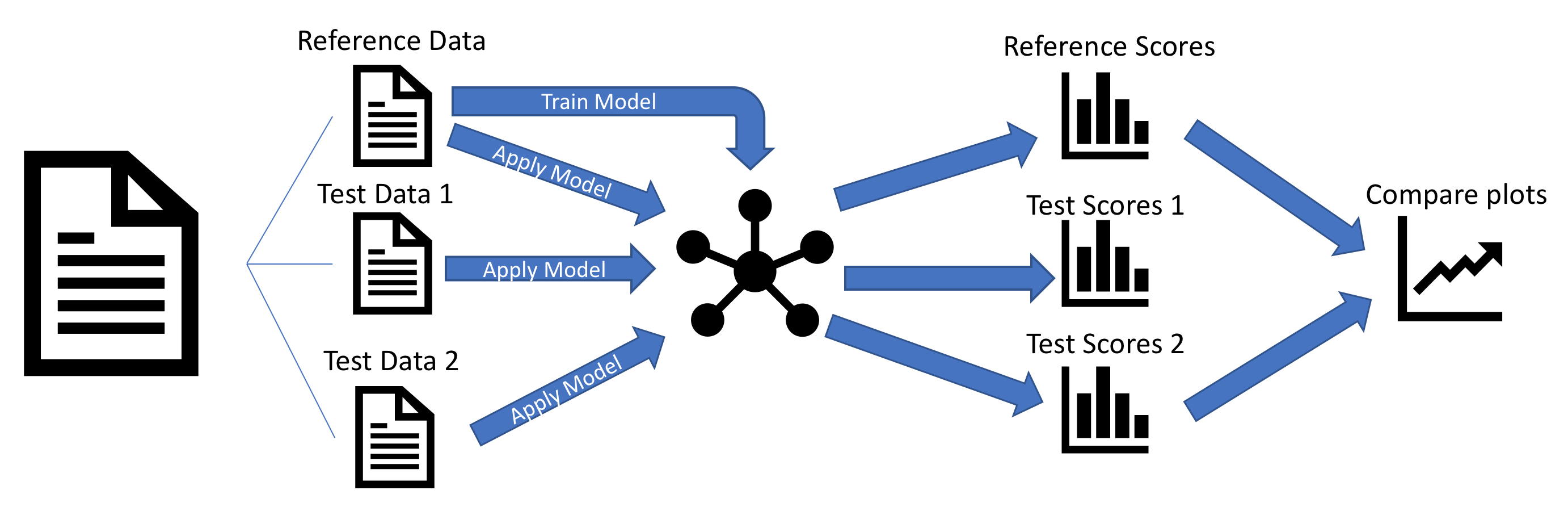}
\caption{General workflow for describing changes between segments of the log file.}
\label{fig:general_workflow}
\end{figure}

Figure \ref{fig:workflow_graphs} shows the relations of the three different plots we introduced. The trace-score plot and drift plot are used to divide the log file into different segments where each segment contains traces with similar behavior. This is done using the temporal aspect of the log file and our sliding window technique. To better understand the differences and equalities among the segments we first take a reference segment (or subset of a segment) that is used to train the model. This model represents the expected traces in the log file based on the reference segment. It can then be applied to every other segment (including the reference segment) to get the partial scores for the traces in that segments. These partial scores can be plotted on an attribute-density plot, making it possible to find the root cause for the difference between the segments. This general workflow for determining the root cause is shown in Figure \ref{fig:general_workflow}. All of the analysis performed in Section \ref{sec:results} are based on this general workflow.

\newpage
\section{Analyzing the data}
\label{sec:results}
In this section we analyze the BPI Challenge 2018 data to show the effectiveness of our proposed workflow. And show that it is able to find interesting insights in the given data.

\subsection{Preprocessing of the data}
Although our method does not require any preprocessing we perform a little bit of preprocessing to avoid overly complex models without any added value over more simple models. We perform the following steps:
\begin{enumerate}
	\item Since our implementation works with CSV-files we first use the Prom Tool \cite{van2005prom} to convert the given XES-files.
	\item We filter out all the \emph{derived attributes} as we are only interested in the attributes that really depend on the activity and not the outcome of the process.
	\item Next we filter out the attributes \emph{eventid}, \emph{event\_identify\_id} and \emph{identity\_id}, since their respective values are unique for every event and would thus form Functional Dependencies with every other attribute in the event.
	\item Because we are primarily looking to test our model for concept drift detection, we also omit the \emph{year} attribute as the process changes every year and we do not want our model to be biased towards this attribute.
\end{enumerate}

All of these preprocessing steps were repeated for every experiment conducted on the data, unless explicitly stated otherwise. Important to note is that our preprocessing steps did not alter the nature of the log or any of the processes in the log.

\subsection{Finding drift points}
\begin{figure}[t]
	\centering
	\includegraphics[scale=0.4]{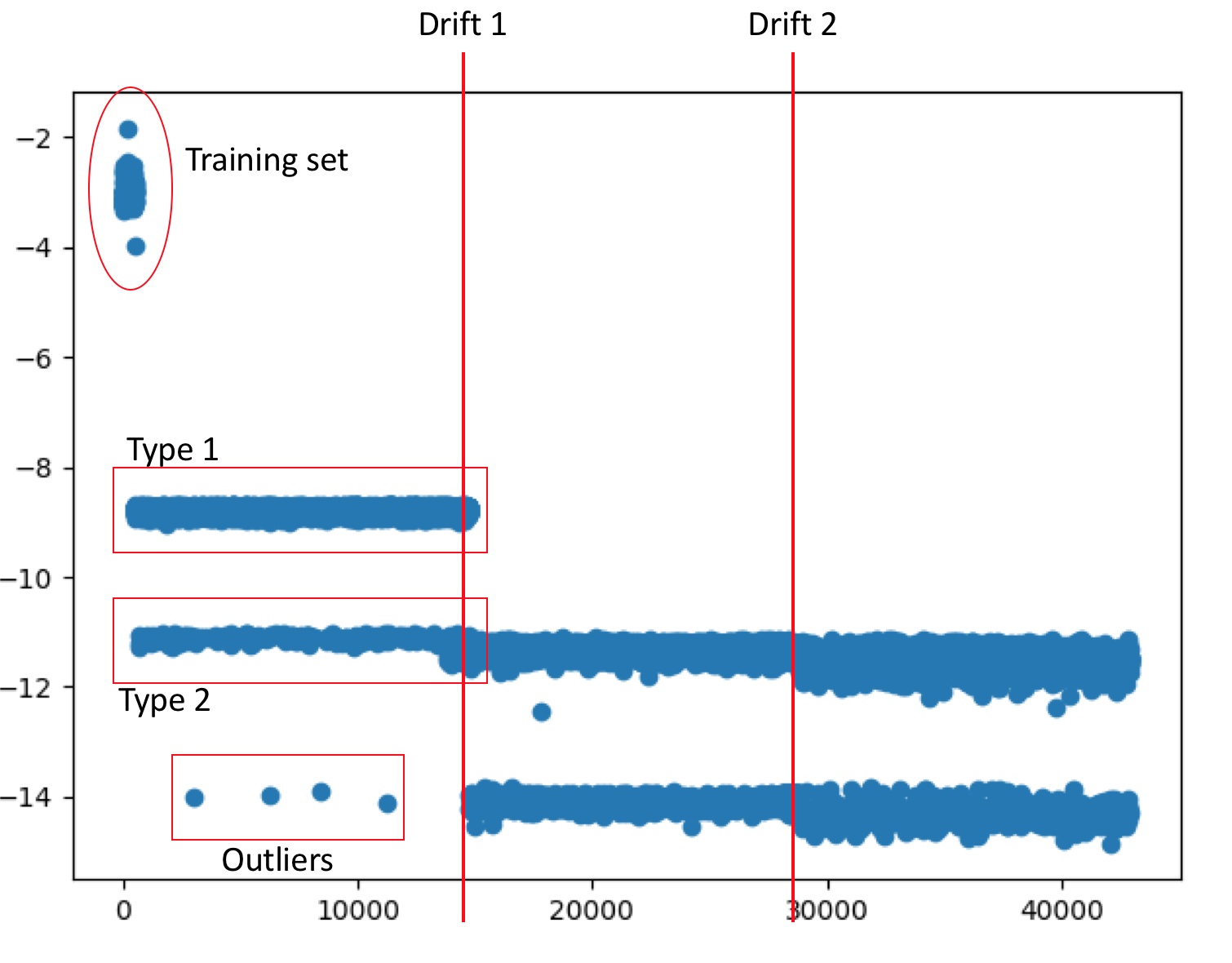}
	\caption{Annotated trace-score plot.}
	\label{fig:scores}
\end{figure}

\begin{figure}
	\centering
	\begin{subfigure}[b]{0.60\textwidth}
		\includegraphics[width=\textwidth]{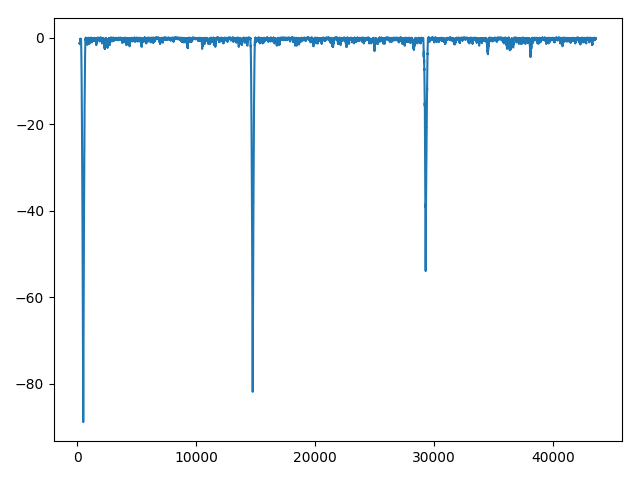}
		\caption{Windows size of 400}
	\end{subfigure}
	\begin{subfigure}[b]{0.60\textwidth}
		\includegraphics[width=\textwidth]{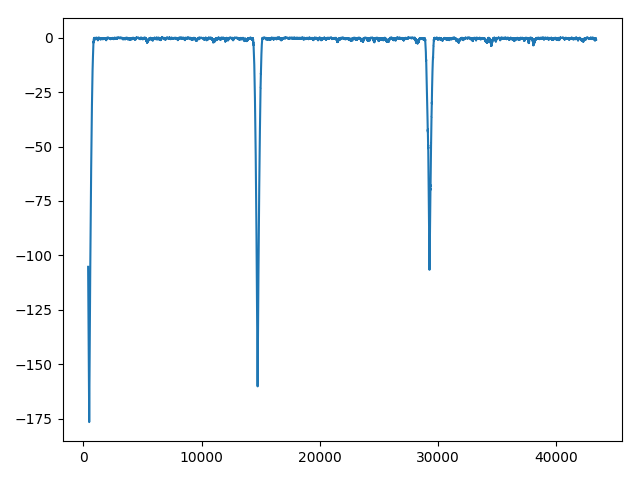}
		\caption{Windows size of 800}
	\end{subfigure}
	\begin{subfigure}[b]{0.60\textwidth}
		\includegraphics[width=\textwidth]{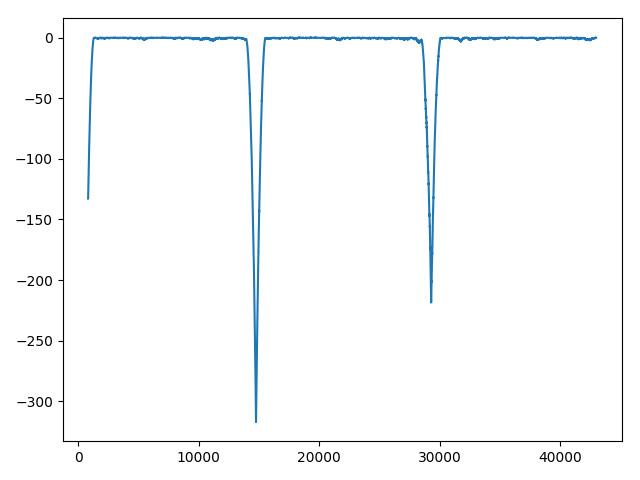}
		\caption{Windows size of 1600}
	\end{subfigure}
	\caption{Drift plots containing the p-values for different window sizes.}
	\label{fig:pvals}
\end{figure}

To look for any drift points in the data we first learn a model according to the log and use this to test all the traces in the following way:
\begin{enumerate}
	\item As a first step we read the data and perform the preprocessing step.
	\item To learn our model we use the first 30,000 events from the log as the training dataset. We use the same dataset for training the model structure as for training all the parameters.
	\item For testing we use all the preprocessed data, including the training set.
	\item We calculate the score for every event in the log given the model.
	\item We calculate the mean score for every trace. This is the \emph{trace score} that we use to further visualize the drifts in the data.
	\item As the first visual step we plot all the scores from the different traces in a trace-score plot, according to the timestamp of the first event of every trace. This is shown in Figure \ref{fig:scores}.
	\item Looking at this plot we can already get some insights about the data:
	\begin{itemize}
		\item The first cluster of traces indicates the traces used for training the model (we can ignore this cluster).
		\item We can clearly see a difference in scores around trace 14,000. Which is a clear indication for a drift point.
		\item A second possible drift point can be seen around trace 29,000, using the drift log in the next step will make it possible to really determine if this is a drift point.
		\item The traces cluster together in two distinct groups. Indicating that the data has mainly two types of traces, each with their own characteristics.
		\item We can see 4 outlying traces before the first drift point. 
	\end{itemize}
	These insights are visually highlighted in Figure \ref{fig:scores} and further explored in the next subsections.
	\item Next we perform the Kolmogorov-Smirnov test with a sliding window over the scores. The resulting drift plots for different window sizes can be found in Figure \ref{fig:pvals}. These plots clearly show the 2 (3 including the training data) drifts occurring in the data, confirming our initial thoughts.
\end{enumerate}

We show in the next subsections that our model is able to explore our findings even further in order to give an explanation for the observed behavior.

\newpage
\subsection{Inspecting the different attributes}
\begin{figure}
	\centering
	\begin{subfigure}[b]{0.6\textwidth}
		\includegraphics[width=\textwidth]{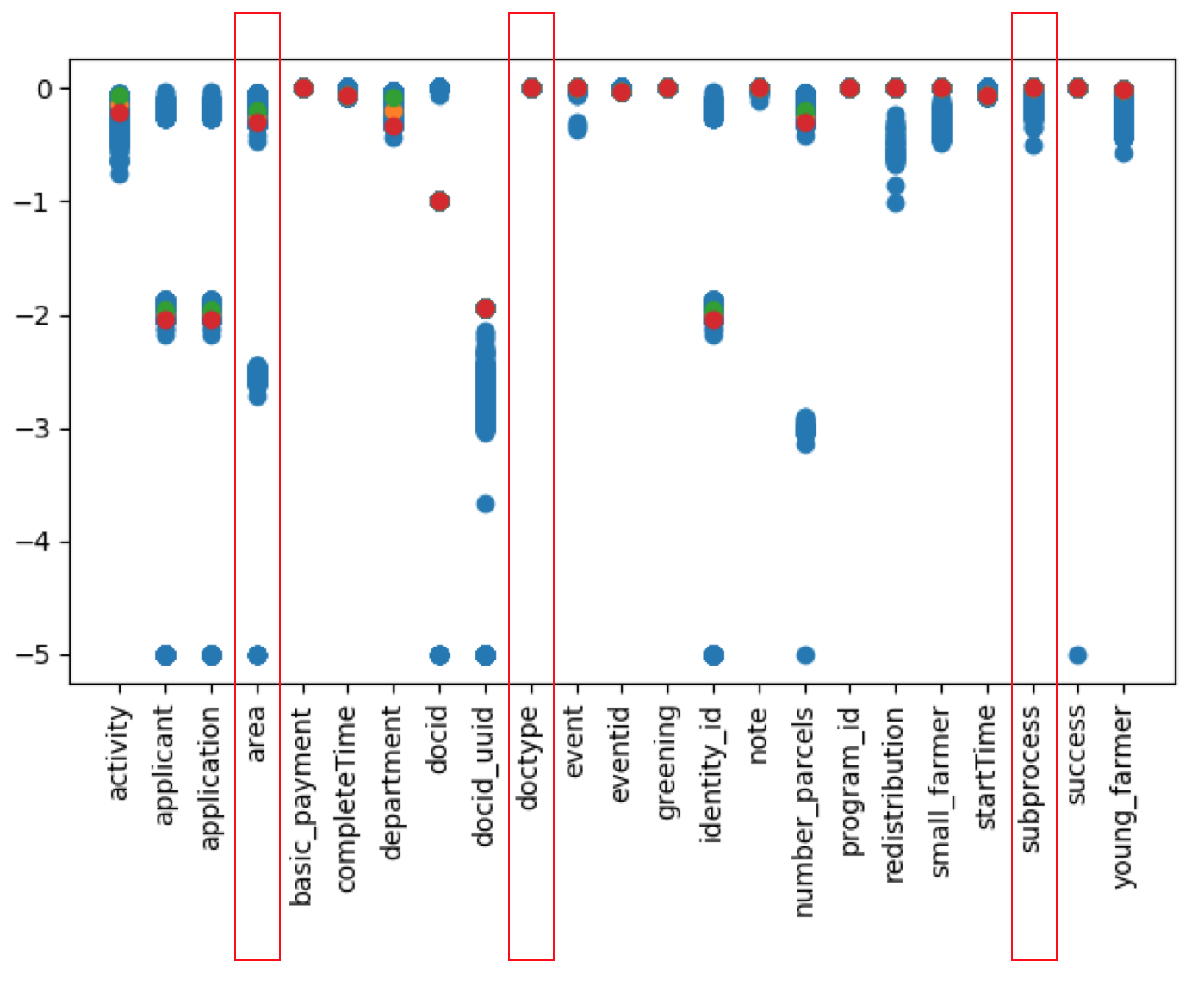}
		\caption{Year 1}
	\end{subfigure}
	\begin{subfigure}[b]{0.6\textwidth}
		\includegraphics[width=\textwidth]{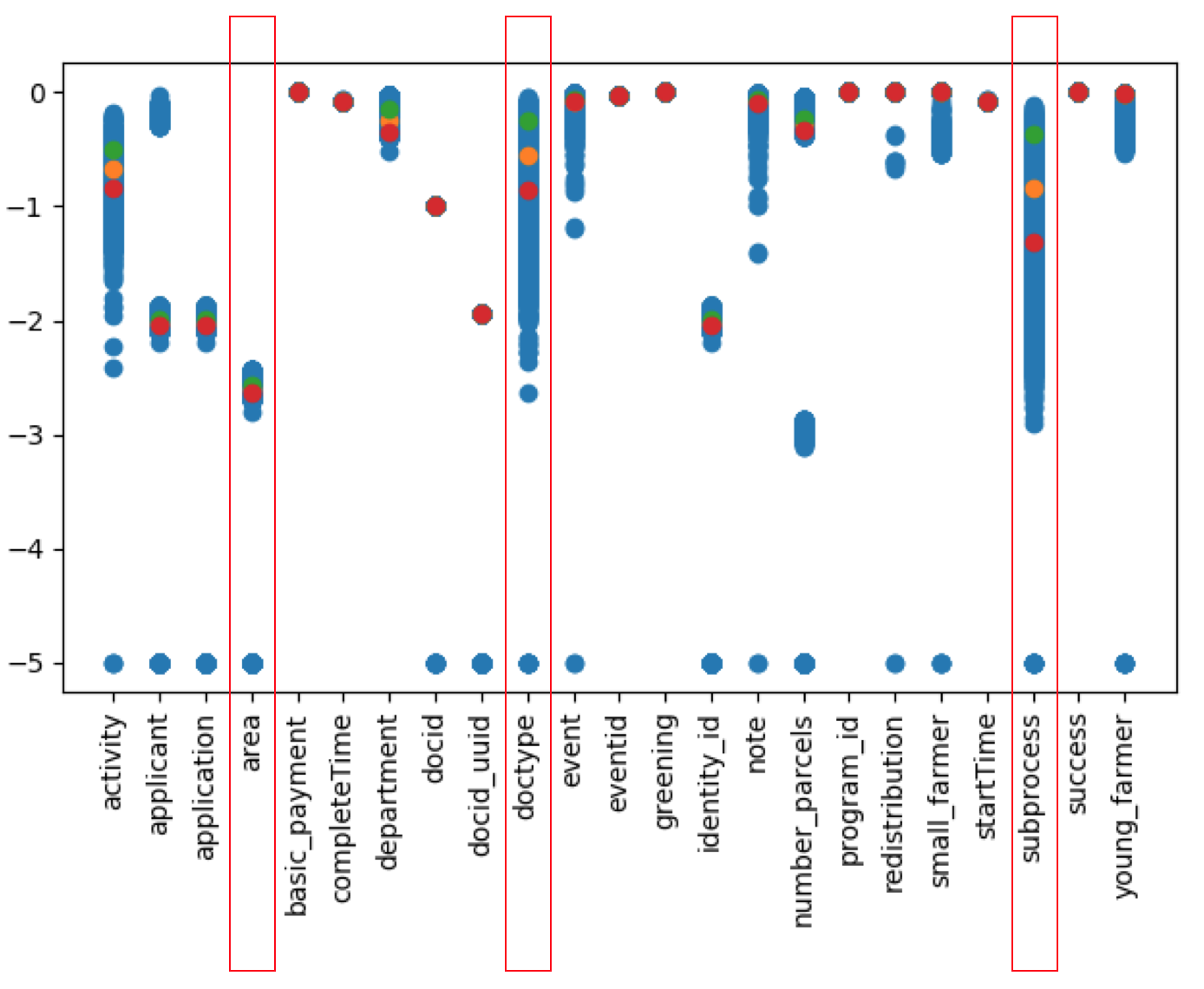}
		\caption{Year 2}
	\end{subfigure}
	\begin{subfigure}[b]{0.6\textwidth}
		\includegraphics[width=\textwidth]{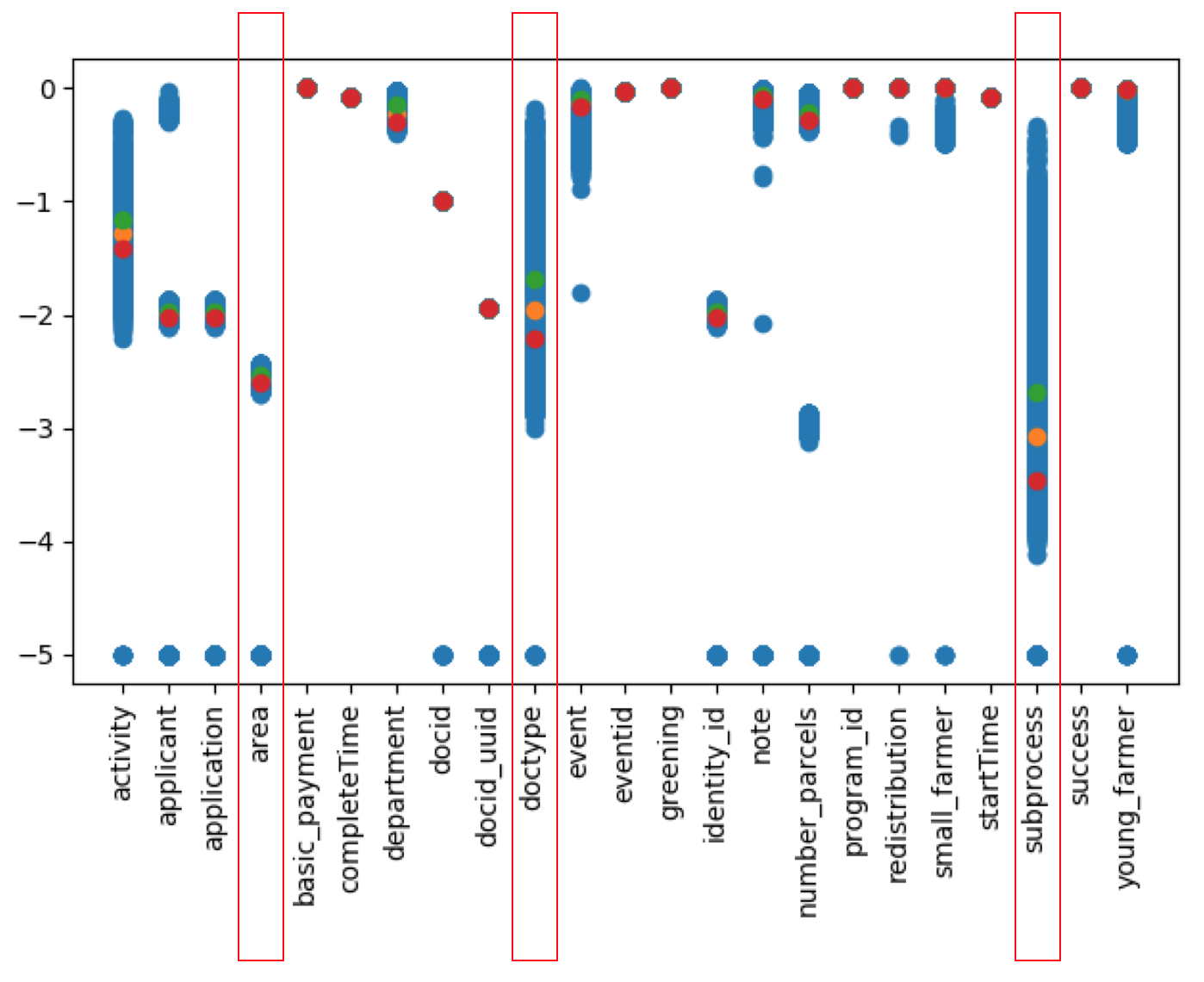}
		\caption{Year 3}
	\end{subfigure}
	\caption{The attribute-density plots for the different years, where we indicated the most changing attributes.}
	\label{fig:attributes_year}
\end{figure}

\begin{figure}
	\centering
	\includegraphics[scale=0.4]{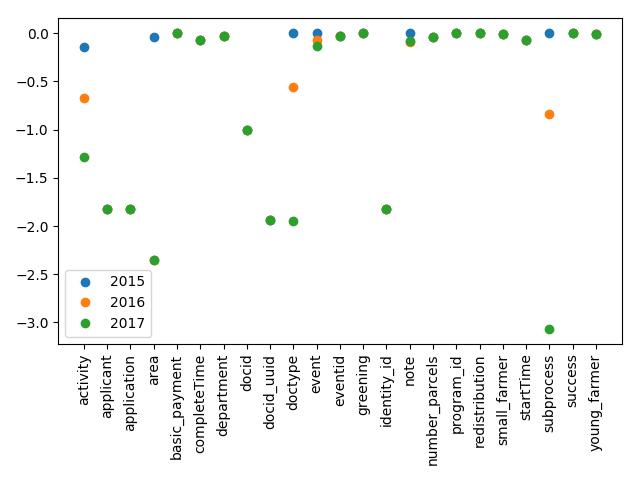}
	\caption{Plot showing all the median values for every year.}
	\label{fig:median_plot}
\end{figure}

In order to find the root cause of the different drifts in the log we use the attribute-density plot to visually represent the characteristics of a process. In order to do so we perform the following steps:

\begin{enumerate}
	\item Learn the model as before.
	\item From every event get the partial scores of every attribute and combine these for the different attributes (in contrast to accumulating the total score for an event as we did before).
	\item Using these partial scores we create multiple attribute-density plots, one plot for every period between drifts. These plots can be used to compare the different segments with each other to determine the root cause for the drifts.
	\item Figure \ref{fig:attributes_year} shows the attribute-density plots for the three different years.
	\item In order to better compare these plots we plotted all the median values in one plot. This plot can be found in Figure \ref{fig:median_plot}.
\end{enumerate}

Using the plots in Figure \ref{fig:attributes_year} and \ref{fig:median_plot} we can see three big changes between the first year, and and the other two years. 
\begin{itemize}
	\item The first difference is found for the attribute \emph{area}, this can be explained by the fact that farmers acquire new land or sell parts of there lands. It is also possible that the used technique for discretizing influences this attribute, as the different bins can change over the years.
	\item The following two changes are probably closely related to each other. Starting in the second year, a new document type was used for the applications, this explains the large differences for this attribute between the different years. And it is normal that due to the changes to the documents, the subprocesses also change. 
	\item Between year 2 and 3 we have the largest difference in the subprocess attribute but less in the other attributes, indicating only some changes in the different subprocesses have occurred.
\end{itemize}

We have thus found two drift points: one at the beginning of each new year. The difference between year 1 and 2 was large as there is a significant change in the area but most importantly a new document type was used that had its consequences on the used subprocesses. It is also possible that this new document type influenced the way the area had to be filled in.
Between year 2 and 3 the difference is less large, this was already visible in the trace-score plot. The most important change between these years was the way the subprocesses where used.

In order to find a more detailed analysis of why this attribute changed that much, other tools and methods can be used to give more detailed explanations and insights.

\newpage
\subsection{Comparing Departments with each other}
\begin{figure}
	\centering
	\begin{subfigure}[b]{0.49\textwidth}
		\includegraphics[width=\textwidth]{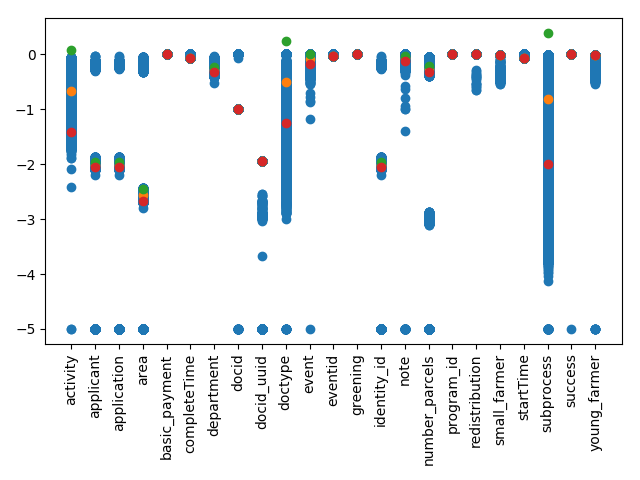}
		\caption{Department 1}
	\end{subfigure}
	\begin{subfigure}[b]{0.49\textwidth}
		\includegraphics[width=\textwidth]{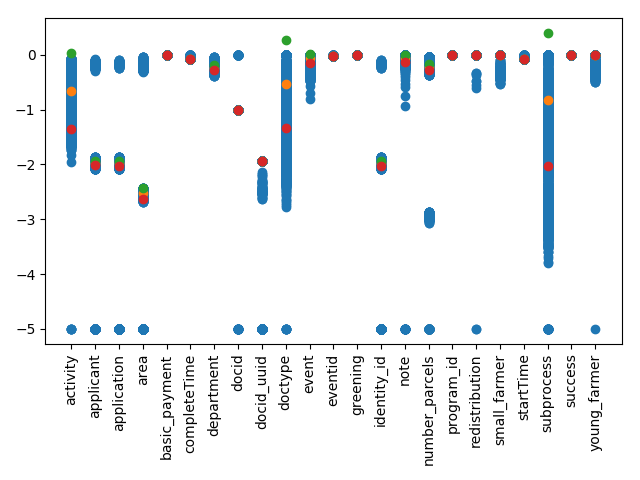}
		\caption{Department 2}
	\end{subfigure}
	\begin{subfigure}[b]{0.49\textwidth}
		\includegraphics[width=\textwidth]{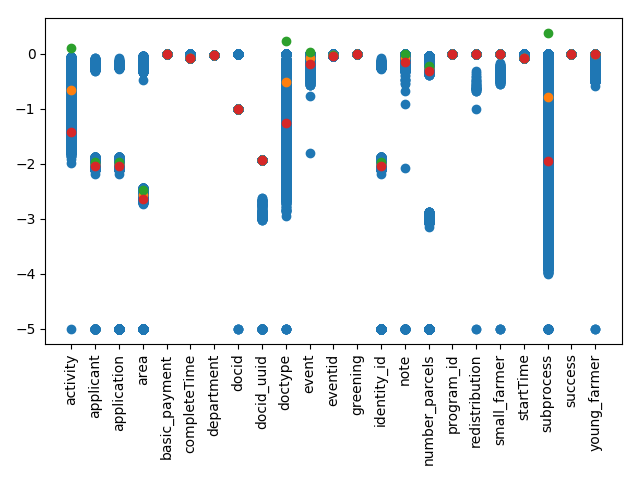}
		\caption{Department 3}
	\end{subfigure}
	\begin{subfigure}[b]{0.49\textwidth}
		\includegraphics[width=\textwidth]{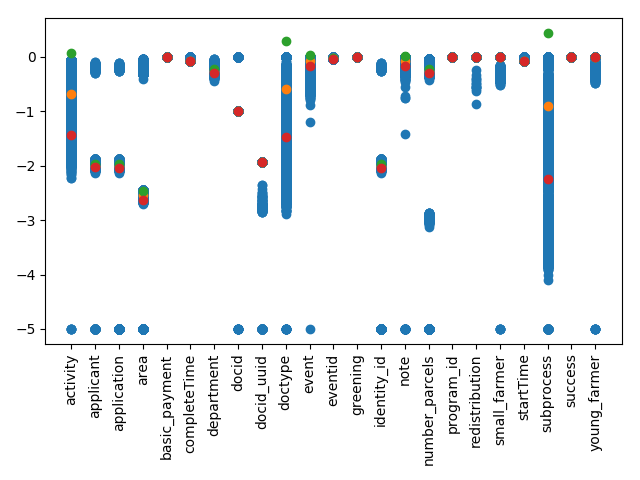}
		\caption{Department 4}
	\end{subfigure}
	\caption{Attribute-density plots for the different departments. When using the first 30,000 events as training data.}
	\label{fig:departments}
\end{figure}

The method we used in the previous sections can be generalized in order to look for differences between groups of traces. We now take a look at possible differences between departments. As a first experiment we use the same training data as before and plot an attribute-density plot for every single department. These plots can be found in Figure \ref{fig:departments}. When comparing these departments with each other we cannot see a clear difference in the scores for the attributes.

\begin{figure}
	\centering
	\begin{subfigure}[b]{0.6\textwidth}
		\includegraphics[width=\textwidth]{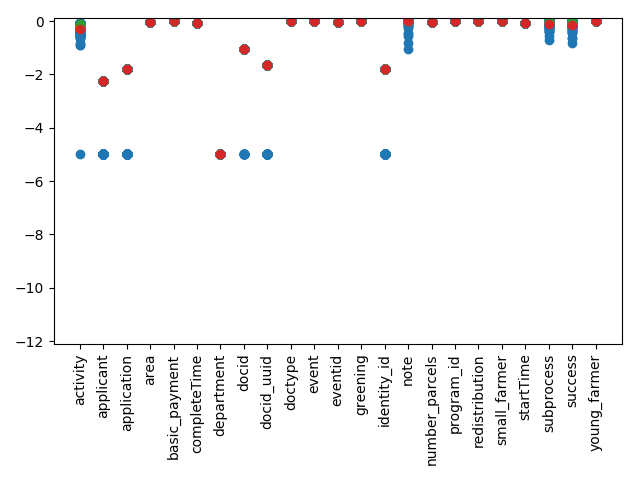}
		\caption{Department 2}
	\end{subfigure}
	\begin{subfigure}[b]{0.6\textwidth}
		\includegraphics[width=\textwidth]{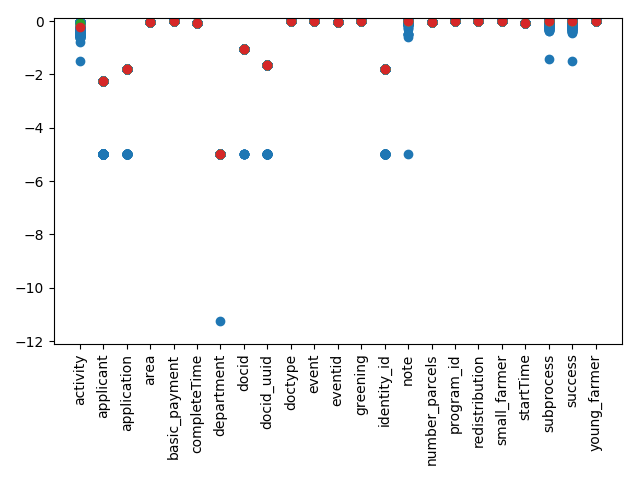}
		\caption{Department 3}
	\end{subfigure}
	\begin{subfigure}[b]{0.6\textwidth}
		\includegraphics[width=\textwidth]{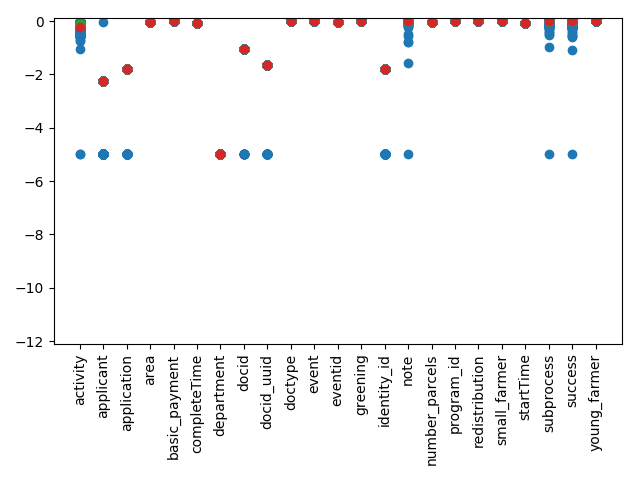}
		\caption{Department 4}
	\end{subfigure}
	\caption{Attribute-density plots after training with department 1.}
	\label{fig:departments_comp}
\end{figure}

In order to better test differences between departments we train our model with events from department 1 and use the other departments as separate test segments. This approach can be generalized for testing any kind of differences. We take a reference dataset for training and use the resulting model for testing other segments of traces. As before we show the attribute-density plots created for every departments in Figure \ref{fig:departments_comp}. We were looking for differences in activities and subprocesses, but when comparing the plots we cannot see a substantial difference between any attribute.

\subsection{Difference between types of traces}
\begin{figure}[t]
	\centering
	\begin{subfigure}[b]{0.49\textwidth}
		\includegraphics[width=\textwidth]{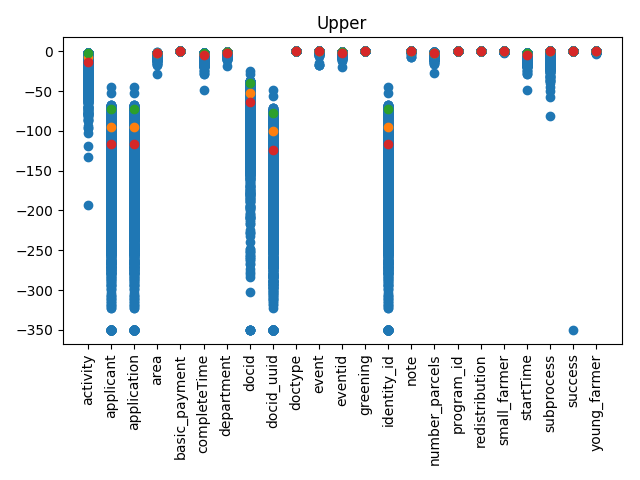}
		\caption{Traces of type 1}
	\end{subfigure}
	\begin{subfigure}[b]{0.49\textwidth}
		\includegraphics[width=\textwidth]{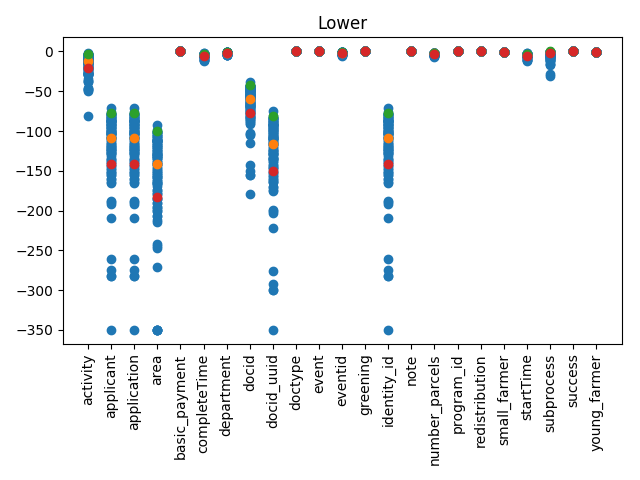}
		\caption{Traces of type 2}
	\end{subfigure}
	\caption{Attribute-density plot for both types of traces}
	\label{fig:area-attribute-graph}
\end{figure}

In the trace-score plot we saw that within every year we have two types of traces. In this section we will take a closer look by decomposing the score even further. We only investigate traces that occur in the first year, as otherwise other drifts and differences might interfere. The first step is to divide the traces in two parts: we have the first type (score between -8 en -10) and the second type (score between -10 and -12). These two types are also highlighted in Figure \ref{fig:scores}. To find a possible explanation for this difference we create an attribute-density plots for both types of traces. These plots can be found in Figure \ref{fig:area-attribute-graph}.

\begin{figure}
	\begin{subfigure}[b]{0.49\textwidth}
		\includegraphics[width=\textwidth]{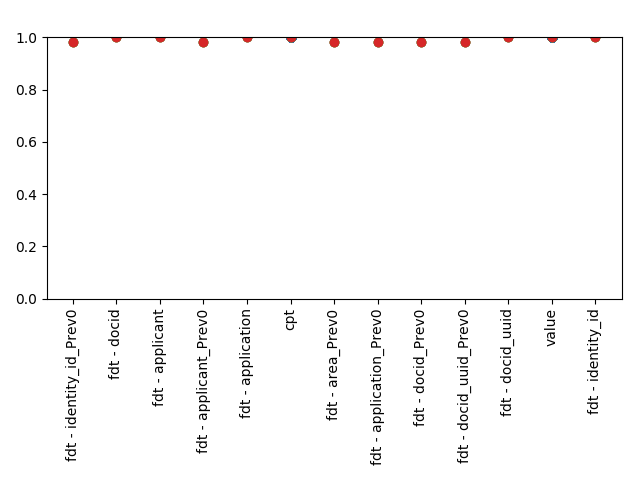}
		\caption{Detailed score for attribute area for type 1 traces}
	\end{subfigure}
	\begin{subfigure}[b]{0.49\textwidth}
		\includegraphics[width=\textwidth]{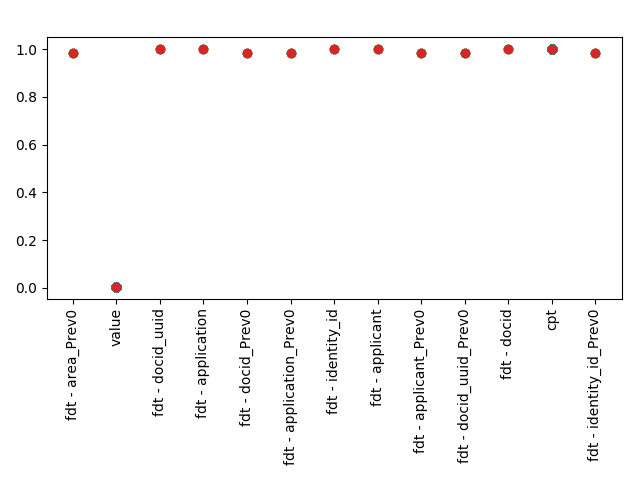}
		\caption{Detailed score for attribute area for type 2 traces}
	\end{subfigure}
	\caption{Detailed attribute-density plot for the attribute area for both types of traces.}
	\label{fig:area-detailed-attribute-graph}
\end{figure}

When comparing these two plots we can see that there is a big difference in the scores for the \emph{area} attribute, which causes the split in two clusters. The score for the attribute \emph{area} is further decomposable in different partial scores: one score is related to the value of the attribute itself. If it is already encountered in the data it returns 1, otherwise it returns the probability of encountering a new value for that attribute based on the training data. The next partial score we got from the Conditional Probability Table, but since the attribute has no Conditional Dependencies this returns 1. Last we get a score for every single Functional Dependency. The results of all these scores can be found in Figure \ref{fig:area-detailed-attribute-graph} for both clusters.

The comparison between these two plots shows that the difference between the two types is caused by the fact that more unseen values for area are encountered in traces from type 2. Thanks to these plots we have a clear insight in which aspect of which attribute causes these deviations. Now we only have to analyze the area attribute into more detail without having to check all attributes present in the log. This effect is, again, due to the way the attribute area was discretized. In the first year we have 148 different values for the area attribute. Of these 148 values only 2 of these values did not appear in the training data, and causes the effect of the two types of traces. Since it is a result from how the data was initially preprocessed we cannot make any conclusions about the real process.

%When examining the data in detail we see that only 473 different values for area are found in the entire log file and 148 values in the first year only. 146 of these values appear in the training set, the other values do not, giving the phenomenon of the two clusters. The log file consists of 15,938 applications from 43,809 applicant. Meaning that multiple applicants must fill in exactly the same area of their properties, when looking at the area value that occurs the most, namely 576.0705 which occurs 15,847 times, we can see that 100 different applicants and applications use this exact size. Since we did not perform any discretization on the data we would not expect this attribute to be so dominant as we intuitively assumed that no two applicants have exact the same area. As we do not have any knowledge on the regulations about these grants we cannot clearly state the cause of these observations, an expert should however be able to figure out why we observed these two types. We can only speculate that this is because of the presence of basic threshold values in the regulation that provide the best chance of getting more money.

\subsection{Investigating anomalous events in the first year}
In the trace-score plot we saw that four dots were not part of any of the two clusters. In this section we use the attribute-density plot to quickly look for any significant differences between the median found in the attribute-density plot for the first year (in Figure \ref{fig:attributes_year}). When comparing these traces to the expected scores we see that especially the \emph{area} and \emph{number\_parcels} attributes have scores that are unexpected when comparing them to the attribute-density plot for the first year. When we investigate these findings we see that these four traces are the only traces in the first year with both the area and number\_parcels equal to 0, thus indeed indicating outlying traces.
\begin{figure}
	\centering
	\includegraphics[scale=0.4]{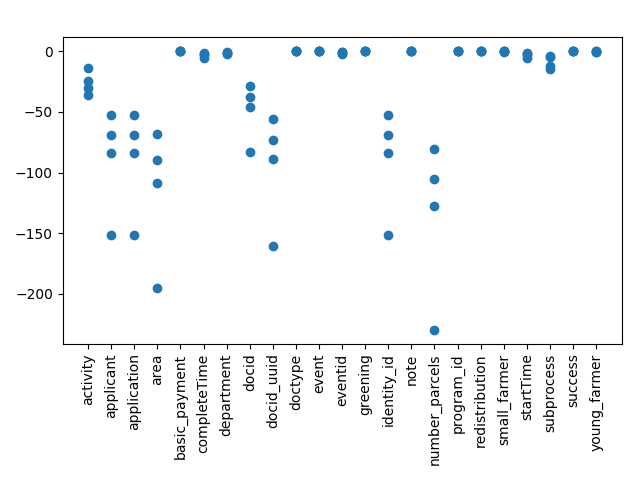}
	\caption{Attribute-densitey plot for the outlier traces}
	\label{fig:anoms}
\end{figure}

\section{Conclusion}
In this report we show that the model we created in order to find anomalies in Business Process logs can easily be used to analyze Concept Drift and other differences in a log file. The way we score events and traces using this model is appropriate for generating more in-depth analysis of why certain traces do deviate. The analysis of the data proved the effectiveness of our method on a real-life dataset where we were able to split our data into the three different years present in the log by only using our drift points.

In this paper we were interested in answering two questions of the BPI Challenge 2018:
\begin{enumerate}
	\item How can one characterize these differences as a particular instantiation of concept drift?
	\item How can one characterize the differences between departments and is there indeed a relation?
\end{enumerate}

To answer this first question we used our trace-score plot and drift plot in order to first detect the drift itself. As showed we did manage to accurately find the drift points in the data. In order to characterize these drifts we further investigated the segments in the log between these drifts. Using the attribute-density plot we were able to detect that, as expected, the document type and subprocesses caused the drifts between the years.

The approach of comparing different years with each other can be generalized in order to solve the second question. We used the first department for training our model and tested the other departments against this model. Besides from the normal deviations in attributes representing ids we did not note any difference between the departments.

As a last analysis we analyzed individual traces and clusters of traces. Again we used the attribute-density plots to find differences between either two clusters or between some events and the expected scores. This method proved to be useful as we were able to successfully explain certain differences in the log file.

The applications mentioned in this report are a starting point but give a clear picture of the possibilities of our eDBNs. The way our analysis was structured should give a insight in the logical steps one can follow to explore multivariate log files. In the future we plan to integrate all these possibilities within a single visual tool. In order to make a quick investigation of these log files more easy. We also plan on incorporating the temporal aspect within traces to be able to detect deviations in duration of a trace.

\bibliographystyle{splncs}
\bibliography{Paper}

\end{document}